\documentclass{article}

\usepackage[utf8]{inputenc}
\usepackage[T1]{fontenc}
\usepackage[margin=1in]{geometry}
\usepackage{graphicx}
\usepackage{booktabs}
\usepackage{xcolor}
\usepackage[numbers]{natbib}
\usepackage{authblk}
\usepackage[hidelinks]{hyperref}

\usepackage{amsmath}
\usepackage{tabularx}
\usepackage{ragged2e}
\usepackage{makecell}
\graphicspath{{../}{./}}

\begin{document}

\title{Asking but Missing: Decision Coverage in Natural-Language Data Querying}

\author[1]{Hangle Hu}
\author[1]{Chenyu Hou}
\author[1]{Bin Cao}
\author[1]{Jing Fan}
\author[1]{Wei Luo}
\affil[1]{Zhejiang University of Technology (zjut.edu.cn)}

\date{}

\maketitle

\begin{abstract}
Large language models are increasingly used to translate natural language questions about structured data into executable programs. A fundamental challenge is recognizing when a question admits multiple computational interpretations and identifying the decisions needed to resolve them. We investigate whether language models recognize these unresolved decisions before generating code, or silently adopt assumptions that may lead to incorrect results. We refer to the latter behavior as Underspecified Decision Completion (UDC). To distinguish different causes of UDC, we introduce Grounded Code Generation (GCG), a diagnostic protocol that separates the identification of unresolved decisions, their resolution using evidence, and their translation into executable code. By examining each stage separately, we can determine where the decisions responsible for incorrect results were first missed or mishandled. We evaluate this framework on BIRD-Pandas with twelve models. The results show that answering clarification questions consistently improves execution accuracy, yet substantial errors remain. Our findings suggest that the main difficulty lies not in answering clarification questions, but in recognizing which questions need to be asked in the first place.
\end{abstract}


\section{Introduction}\label{sec:introduction}

Natural language interfaces allow users to query structured data without writing code. Many of these systems use large language models (LLMs) to translate a question into an executable program~\citep{chen2021codex,li2024bird,hong2025nextgen,hu2026bird}. Consider a question such as ``Which districts had the most withdrawals?'' This could mean the districts with the largest number of withdrawals or those with the highest total amount withdrawn. A language model may choose one interpretation without checking what the user intended. If the resulting program gives the wrong answer, we cannot tell whether the model overlooked the need for clarification, received insufficient information, or made a mistake when writing the code. Much like debugging a program, finding the error requires us to examine the steps that led to the result. We therefore investigate what questions the model asks, what information it receives, and how it uses that information to generate code.

Prior work has studied when LLMs should ask for clarification~\citep{zhang2025modeling}, and how to detect and resolve ambiguity in program generation, including interactive Text-to-SQL methods~\citep{yang2026orchid,zhang2024clamber,mu2024clarifygpt,wu2025clarifycoder,vijayvargiya2025ambigswe,huo2026birdinteract}. Some of these studies measure whether a model detects that a question is ambiguous and evaluate the quality of clarification and interaction. Other methods, including Chain-of-Thought prompting, retrieval augmentation, and implicit reasoning training~\citep{wei2022chain,lewis2020rag,yang2026imprif}, improve how a model uses available knowledge to resolve identified decisions. Error analysis and self-correction start from a program that has already been generated~\citep{ashury2026erroratlas,jiang2026reflexicoder}. Repeated generations provide limited information about an open decision, because a model can return the same interpretation across samples~\citep{richter2026collapse}. Metrics such as Pass@k, BLEU~\citep{papineni2002bleu}, and CodeBLEU~\citep{ren2020codebleu} score the generated program and give the same outcome whether the failure comes from an unasked decision, insufficient evidence, or an incorrect implementation. A model must recognize that a decision is open before it can ask about it. An earlier question therefore remains important for understanding the limits of these approaches: before generating a program, can a model raise clarification questions for the execution decisions that the input prompt leaves open?

\begin{figure}[t!]
\centering
\includegraphics[width=0.99\linewidth]{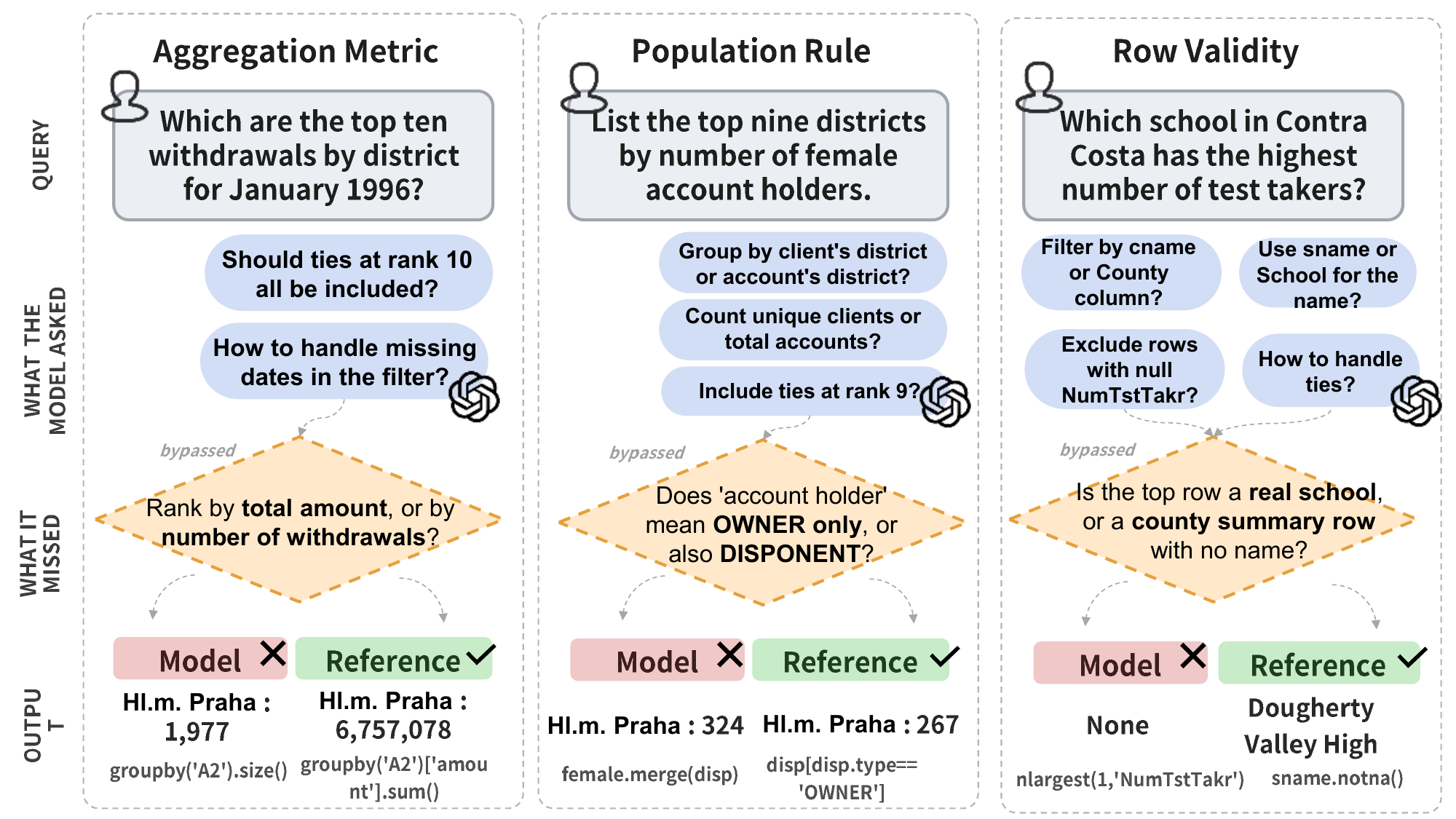}
\caption{Three examples of UDC, in which a model asks several clarification questions and receives answers, yet does not ask which quantity to rank by, who counts as an account holder, or whether a summary row is a valid entity, so its program returns a result that differs from the reference.}
\label{fig:hypothesis-diagram}
\end{figure}

This question concerns a specific pattern. Figure~\ref{fig:hypothesis-diagram} illustrates three cases. In each case, the model asks relevant clarification questions, for example about how to break ties or handle missing values, and receives correct answers. The decision that determines the result is different from the decisions those questions address: the quantity that ``top'' ranks, who counts as an account holder, or whether a summary row is a valid entity. The model completes each of these decisions with a default assumption and returns an incorrect result. We call this pattern \textbf{Underspecified Decision Completion (UDC)}. An \emph{execution decision} is an implementation choice whose alternatives can produce different outputs on at least one dataset that fits the schema. As in Text-to-SQL benchmarks such as BIRD, where the annotated reference program serves as the proxy for the user's intended result~\citep{li2024bird,hu2026bird}, we define an execution decision as \emph{underspecified} when the reference program uses an alternative that the input prompt does not state explicitly. UDC occurs when the model omits an underspecified decision from its clarification questions before writing a program.

To examine how UDC relates to program errors under the benchmark evaluation, we predict that errors will remain after the clarification questions a model raises have been answered, and that some of these errors involve execution decisions in the reference program that the model did not raise. Testing this requires locating each execution decision in a program, so we use NL-to-Pandas, the task of generating pandas programs from natural language questions. In SQL, a query specifies the intended result and the database engine determines how to evaluate it~\citep{wu2019dataframe,hagedorn2021pandas}. A pandas program applies a sequence of operations to the data~\citep{wu2019dataframe,hagedorn2021pandas,petersohn2020scalable}, so a decision such as how to rank or filter rows appears as a specific operation that can be compared with the reference program. We propose Grounded Code Generation (GCG), a diagnostic protocol with three phases: question raising, evidence grounding, and program synthesis. In the first phase, the model lists the execution decisions it regards as open and states each as a clarification question, before any program is written. We answer each question using the reference program, and the model then writes the program from the input prompt, its questions, and these answers. The answers describe the decisions in plain language and contain no reference code. For each program that remains incorrect, GCG locates the first phase at which the decision that determines the output is missing.

We apply this protocol to twelve models on the BIRD-Pandas benchmark~\citep{hu2026bird}. Answering the clarification questions that models raise improves execution accuracy for all twelve models, but errors remain. For four models analyzed in depth, the largest share of the remaining errors is traced to execution decisions in the reference program that the model did not ask about, rather than to insufficient evidence or to incorrect program synthesis. On the failing prompts, the clarification questions the models raise address most of the execution decisions they identify as open, so the gap lies in how many of the relevant decisions the questions cover. Many of the decisions that go unasked are not stated explicitly in the input prompt. In a system that returns only the answer, the decisions the model completed on its own are not visible from the output.

Our contributions are as follows. First, we identify and characterize Underspecified Decision Completion (UDC), a failure in which a model does not ask about an execution decision left unspecified by the input but resolved in the benchmark reference program. This allows us to examine whether models recognize the decisions necessary to produce the expected result, beyond whether their generated programs execute correctly. Second, we introduce Grounded Code Generation (GCG), a diagnostic protocol for investigating where these failures arise. By separating the questions a model asks, the evidence it receives, and the program it generates, GCG allows us to distinguish overlooked decisions from errors in evidence grounding and code generation. Third, we conduct an empirical study of twelve language models on BIRD-Pandas. Although answering clarification questions improves execution accuracy across all models, our detailed analysis of four models finds that overlooked decisions account for the largest share of remaining failures. Most of these omissions concern decisions not explicitly specified in the input, suggesting that recognizing what needs clarification remains a major difficulty.

\section{Related Work}

\paragraph{Code Generation and Evaluation.}
Large language models trained on code~\citep{chen2021codex,roziere2023code} and specialized Text-to-SQL parsers~\citep{wang2020rat-sql,scholak2021picard} have made natural language interfaces to structured data increasingly practical, and benchmarks for SQL and pandas now measure this ability on realistic databases~\citep{yu2018spider,li2024bird,hu2026bird}. Reasoning and retrieval techniques such as Chain-of-Thought prompting, retrieval augmentation, and implicit reasoning training~\citep{wei2022chain,lewis2020rag,yang2026imprif} aim to improve how models use the knowledge available to them. Evaluation in this line of work relies on Pass@k, exact match, and execution accuracy, and benchmarks such as BIRD and BIRD-Pandas take the annotated reference program as the proxy for the user's intended result~\citep{li2024bird,hu2026bird}. These measures indicate whether the final program is correct, but they do not distinguish which step in the process introduced an error or whether the model identified the decision that determines the output.

\paragraph{Underspecification and Clarification.}
A growing body of work addresses ambiguous or underspecified requirements in program generation. Some studies measure how ambiguity affects model performance and how well models identify it~\citep{yang2026orchid,zhang2024clamber,hu2026bird}. Other work lets models ask clarification questions or choose among interpretations before generating a program~\citep{mu2024clarifygpt,wu2025clarifycoder,cao2025pragmatic,vijayvargiya2025ambigswe,clareval2025,fang2026clarifycodebench}, or repair the ambiguous description itself~\citep{jia2025specfix}. In Text-to-SQL, related systems detect ambiguity and resolve it through interaction~\citep{huo2026birdinteract,zhao2024sphinteract,ding2025ambisql,zhang2025clear}. These studies evaluate final task success, the quality of clarification, or the quality of interaction. A related pipeline in~\citep{hu2026bird} shares the three-phase structure of raising questions, grounding evidence, and synthesizing a program, but aims to complete a task rather than to diagnose where an error originated. Error analysis for structured data queries is closer to our approach. Work on NL2SQL categorizes semantic errors in generated queries~\citep{liu2025nl2sqlbugs}. As in these studies, we judge whether a deviation matters against the intended result, which here is the benchmark reference. Our focus is different: we measure what proportion of the open execution decisions a model asks about, and we use error categories to organize which decisions should be surfaced before a program is written.

\paragraph{Failure Analysis and Output-Based Detection.}
Failure-oriented research builds cross-task error taxonomies~\citep{ashury2026erroratlas} and studies how models critique, detect, and correct their own errors~\citep{jiang2026reflexicoder,cho2025cocos,phan2026hidden,kumaran2026confidence}. In Text-to-SQL, error-guided correction repairs a query after its failure has been categorized~\citep{chaturvedi2025sqlofthought}. These approaches analyze or revise an output that has been generated. The failure we study occurs earlier, when the decision that determines the output is not identified before the program is written, and our protocol assigns each error to the first phase at which that decision is missing. Detecting such failures from outputs alone is difficult. A model can return the same incorrect interpretation across samples when the input prompt leaves decisions open, so output agreement does not indicate that the input prompt was fully specified~\citep{richter2026collapse}. Instead of relying on output agreement, we evaluate the clarification questions a model raises against the benchmark reference program.

\section{Methodology}\label{sec:method}

\subsection{Task and Execution Decisions}\label{sec:method:preliminaries}

\paragraph{Task.}
Given a natural language question $Q$, schema information $S$ (column names, data types, and representative values), and task-provided external knowledge $K$, the model generates a pandas program $c$. Let $\mathcal{I} = (Q, S, K)$ denote this input, following the standard BIRD-Pandas format. Let $O(c, D)$ denote the output of executing $c$ on the CSV dataset $D$, and let $c^\star$ denote the reference program provided by the benchmark. A program is correct when $O(c, D) = O(c^\star, D)$. Following this convention, our diagnostic protocol also uses $c^\star$ to identify which execution decisions the output depends on.

\paragraph{Execution Decisions and Implicit Assumptions.}
An \emph{execution decision} is an implementation choice that can affect the output: two programs $c_1$ and $c_2$ that differ in this decision satisfy $O(c_1, D') \neq O(c_2, D')$ on at least one dataset $D'$ consistent with $S$. An execution decision is \emph{underspecified} when the input $\mathcal{I}$ does not determine which alternative the model should take. When a decision is underspecified, the model completes it on its own, and the alternative it adopts is its \emph{implicit assumption} for that decision.

\subsection{Deriving Coverage Dimensions from Observed Failures}\label{sec:method:taxonomy}

\paragraph{Open Annotation of Discrepancies.}

The input $\mathcal{I}$ over structured data often leaves execution decisions open, and existing clarification taxonomies typically start from a predefined list of missing information such as ambiguous column or value references. These taxonomies do not cover choices such as result cardinality or tie handling that affect program outputs but are not visible at the schema level. We instead derive our coverage dimensions from the execution decisions that direct generation fails to surface, which helps ensure that the dimensions include execution-critical choices that schema-level taxonomies miss. An annotator model (DeepSeek\-V4\-Pro) compares each generated program with the reference program and records each logical deviation as a separate discrepancy, without receiving a predefined category list. Each discrepancy receives a free-text description and an open \texttt{VERB\_OBJECT} label such as \texttt{OMIT\_TIE\_HANDLING}. For each discrepancy, an \textbf{Explicit Level} records how clearly the input $\mathcal{I}$ ($Q$, $S$, and $K$ together) states the decision: \texttt{EXPLICIT}, \texttt{PARTIAL}, \texttt{IMPLICIT}, or \texttt{NO\_CLUE}, where \texttt{NO\_CLUE} marks that the input contains no textual clue toward the alternative the reference program adopts. An \textbf{Evidence Source} records which of $Q$, $S$, and $K$ contains a clue for the decision, or \texttt{N} when none does. A fixed rule set maps the open label to one of six failure actions and one of seven primary query objects (Appendix~\ref{app:taxonomy:process}). Representative cases and per\-model distributions are in Appendices~\ref{app:taxonomy:instances} and~\ref{app:taxonomy:distributions}. The annotation covers discrepancies from three models (Qwen3-Max, DeepSeek-V4-Pro, and Gemini-3.1-Pro). The pooled distribution is in Table~\ref{tab:method-object-routing}; omissions and substitutions together account for about 76\% of discrepancies.

\paragraph{From Discrepancies to Dimensions.}
The dimensions follow the order in which a program fixes its execution decisions: which data is referenced, which records are included, how they are computed, what is returned, and how boundary cases are handled. Each primary query object is assigned to the dimension it primarily concerns (Table~\ref{tab:method-object-routing}). Appendix~\ref{app:taxonomy:mapping} provides the full mapping rules. A discrepancy can concern multiple execution decisions and therefore map to more than one dimension. For instance, a computation discrepancy may concern the business definition of a measure (\texttt{BUSINESS\_RULE}), the aggregation method (\texttt{EXECUTION\_METHOD}), or the required result count (\texttt{RESULT\_CONTRACT}).

\paragraph{Five Coverage Dimensions.}
The five dimensions are as follows. \texttt{DATA\_REFERENCE} covers which table, column, value, entity, or data source is referenced. \texttt{BUSINESS\_RULE} covers which inclusion, exclusion, status, threshold, time, null eligibility, or metric definition rule applies. \texttt{EXECUTION\_METHOD} covers which relation, aggregation, counting, deduplication, conversion, or computation order realizes the decision. \texttt{RESULT\_CONTRACT} covers what fields, granularity, ordering, row count, or tie policy is required. \texttt{BOUNDARY\_GUARD} covers what empty-set, missing-value, zero-division, no-match, or fallback behavior is required. \texttt{BOUNDARY\_GUARD} is kept separate because boundary behavior can change outputs and is frequently not stated in the input. These dimensions are based on the reference program and serve as a diagnostic checklist for Phase~1; they cover the decisions relevant to this benchmark rather than a general-purpose taxonomy of ambiguity, and they may differ from the questions a model raises without this checklist.

\begin{table}[t!]
\caption{Pooled distribution of the preliminary discrepancies and the Phase~1 dimension each object maps to.}
\label{tab:method-object-routing}
\centering
\footnotesize
\begin{tabularx}{\linewidth}{@{}>{\RaggedRight\arraybackslash}p{0.20\linewidth} >{\Centering\arraybackslash}p{0.10\linewidth} >{\Centering\arraybackslash}p{0.14\linewidth} >{\Centering\arraybackslash}p{0.14\linewidth} X@{}}
\toprule
\textbf{Primary query object} & \textbf{Count} & \textbf{Share} & \textbf{No clue (\texttt{N})} & \textbf{Phase~1 dimension} \\
\midrule
\texttt{COMPUTATION}      & 647 & 23.7\% & 38.8\% & \texttt{EXECUTION\_METHOD} \\
\texttt{SELECTION\_RULE}  & 479 & 17.5\% & 30.7\% & \texttt{BUSINESS\_RULE} \\
\texttt{DATA\_REFERENCE}  & 398 & 14.6\% & 32.7\% & \texttt{DATA\_REFERENCE} \\
\texttt{RESULT\_CONTRACT}  & 374 & 13.7\% & 55.1\% & \texttt{RESULT\_CONTRACT} \\
\texttt{REPRESENTATION}    & 331 & 12.1\% & 23.3\% & \texttt{DATA\_REFERENCE} or \texttt{BUSINESS\_RULE} \\
\texttt{RELATIONSHIP}      & 330 & 12.1\% & 35.8\% & \texttt{EXECUTION\_METHOD} \\
\texttt{BOUNDARY\_GUARD}   & 171 & 6.3\%  & 81.3\% & \texttt{BOUNDARY\_GUARD} \\
\midrule
\textbf{Total}             & 2{,}730 & 100.0\% & 39.1\% & \\
\bottomrule
\end{tabularx}
\end{table}

\subsection{Grounded Code Generation: A Three-Phase Diagnostic Protocol}\label{sec:method:framework}

\paragraph{Protocol Overview.}

\begin{figure*}[t!]
\centering
\includegraphics[width=0.98\textwidth]{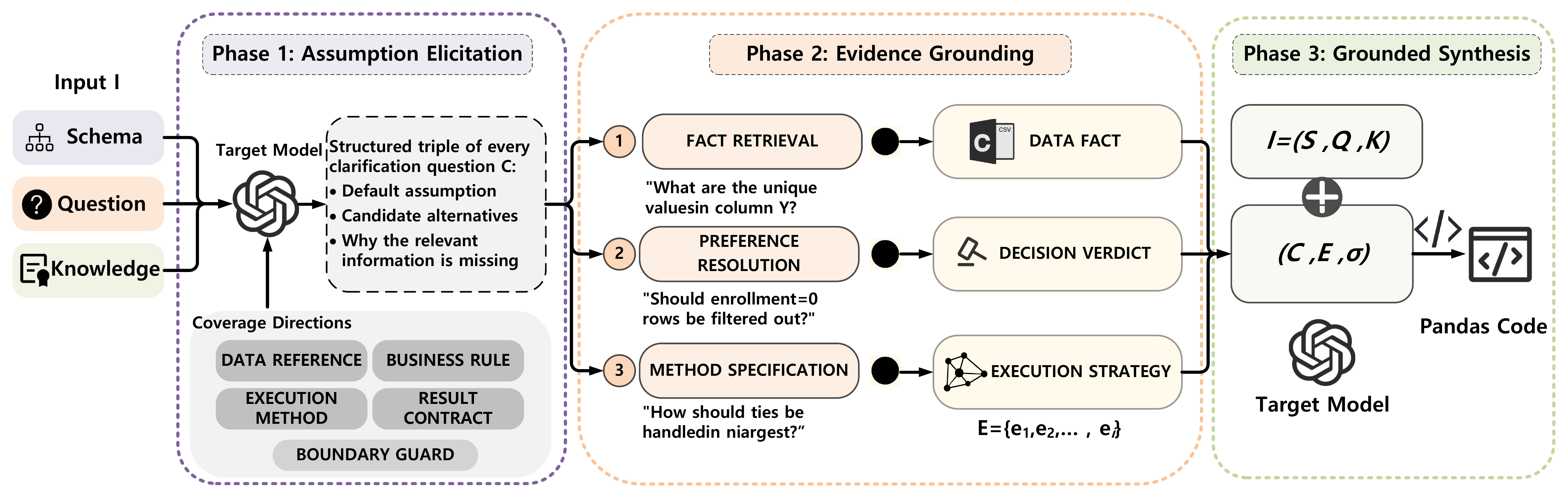}
\caption{Overview of the Grounded Code Generation (GCG) framework.}
\label{fig:gcg-architecture}
\end{figure*}

In direct generation, identifying a decision, obtaining evidence, and implementing it all occur inside a single output, which means that the point at which a decision is lost is not visible. GCG separates these steps (Figure~\ref{fig:gcg-architecture}). Given the input $\mathcal{I}$, Phase~1 lists the execution decisions the model regards as unresolved and turns them into clarification questions, without generating a program. Phase~2 answers those questions with decision-level evidence. Phase~3 generates the final pandas program from the input, the questions, and the evidence. Each phase leaves a record, so a remaining failure can be traced to the first phase at which the decision that determines the output was lost. Abridged prompts for each phase are in Appendix~\ref{app:framework}, and Appendix~\ref{app:example-q22} walks through one example across all three phases.

\paragraph{Phase~1: Assumption Elicitation.}

Phase~1 receives the input $\mathcal{I}$ and does not generate a program. For each execution decision it identifies as unresolved, the model states the assumption it would otherwise adopt, lists the alternatives, and explains why $\mathcal{I}$ does not determine among them, before turning the gap into a clarification question. This happens before any program is written, so the model's default choice is visible rather than embedded in the output. Each clarification question addresses one of five dimensions, and the prompt asks a concrete question for each dimension: \emph{which} column, rule, or method is intended, rather than a general question about whether the input is ambiguous.

The Phase~1 prompt directs the model to inspect the five coverage dimensions, so that it raises questions on execution decisions across all dimensions. The dimensions guide inspection; the model returns free-form gap descriptions and clarification questions without emitting dimension labels. Each question takes one of three surface forms: a \emph{data query}, a \emph{rule verification}, or a \emph{method clarification}. We denote the resulting set of questions by $\mathcal{C}$.

\paragraph{Phase~2: Evidence Grounding.}
Phase~2 receives the clarification questions $\mathcal{C}$. A question may address several assumptions, so Phase~2 first decomposes each one into atomic sub-questions and answers each one separately. Each sub-question is assigned an evidence type according to its surface form. Data queries are answered directly from the CSV dataset $D$; rule and method questions are answered by an oracle that may consult the reference program $c^\star$:
\begin{itemize}
  \item \texttt{FACT\_RETRIEVAL} (from a \emph{data query}): a targeted query over the CSV data returns the requested data fact, an observable property of $D$;
  \item \texttt{PREFERENCE\_RESOLUTION} (from a \emph{rule verification}): the oracle returns a YES/NO judgment on the proposed rule together with the rule that applies;
  \item \texttt{METHOD\_SPECIFICATION} (from a \emph{method clarification}): the oracle returns an execution strategy, that is, the required operation or parameter setting.
\end{itemize}
Let $E$ denote the resulting evidence set. Each item records its sub-question, evidence type, returned answer, and evidence source in $\{\texttt{Q},\texttt{S},\texttt{K},\texttt{N}\}$, i.e., the field of $\mathcal{I}$ that supports the answer, or $\texttt{N}$ when none does. $E$ contains only evidence requested by the clarification questions raised in Phase~1. Queries over $D$ are deterministic, so \texttt{FACT\_RETRIEVAL} answers are reproducible.

\paragraph{Oracle Conditioning as a Control.}

The oracle returns decision\-level answers in plain language, and Phase~3 receives only these answers. By design, each raised clarification question receives an answer consistent with the reference program, so the protocol focuses on what remains unresolved once the questions are answered. Appendix~\ref{app:validation:oracle} reports how closely the evidence follows this design.

\paragraph{Phase~3: Grounded Synthesis.}
Phase~3 receives
\[
\mathcal{H}=(\mathcal{I},\mathcal{C},E,\sigma),
\]
where $\sigma$ is an optional dataset instruction that controls only the output format. The model generates the final pandas program from $\mathcal{H}$: $E$ answers the questions in $\mathcal{C}$, and $K$ remains available for anything Phase~1 did not raise. When $E=\emptyset$, Phase~3 generates from $\mathcal{I}$ and $\sigma$ alone, which is equivalent to direct generation.

\subsection{Failure Attribution}\label{sec:method:attribution}

\paragraph{Error Attribution by Phase.}
For each input that remains incorrect after Phase~3, we assign one label in the following priority order, corresponding to the first phase at which the decision that determines the output is lost: 
\begin{enumerate}
  \item \texttt{Phase1\_Missed\_Gap}: the decision that determines the output is not semantically covered by the Phase~1 question set;
  \item \texttt{Phase2\_Insufficient}: the decision is raised, but the Phase~2 evidence is insufficient, misaligned, or at the wrong level of detail;
  \item \texttt{Phase3\_Selective\_Neglect}: sufficient evidence is available, but the generated program wholly or partly ignores it;
  \item \texttt{Phase3\_Translation\_Error}: sufficient evidence is available and used, but translated into an incorrect pandas operation.
\end{enumerate}

\paragraph{Priority Rule and Validation.}
Phase~2 answers only the questions that Phase~1 raised, and Phase~3 uses only the evidence it receives. Under this information flow, a decision that does not receive evidence at an earlier phase has no corresponding evidence at a later phase, so we attribute the failure to the earlier phase and do not count it again downstream. Each failure therefore receives one label, at the first phase of loss. When several lost decisions depend on a prerequisite that Phase~1 did not raise, the case is assigned to that Phase~1 loss. Throughout this paper, \texttt{Phase1\_Missed\_Gap} corresponds to a decision lost at Phase~1, \texttt{Phase2\_Insufficient} to loss at Phase~2, and \texttt{Phase3\_*} to loss at Phase~3. A constrained LLM diagnostician assigns these labels from the recorded three-phase trace, and human annotation validates both the Phase~1 coverage judgment and the earliest-phase label (Appendix~\ref{app:validation:coverage}).

\subsection{Phase~1 Coverage Metrics}\label{sec:method:coverage}
To measure how many of the execution decisions in the reference program Phase~1 raises on cases where the program remains incorrect, let $G_i$ denote the decisions identified from the reference program for input $i$, and let $R_i$ denote the decisions covered by the Phase~1 questions. We report the following pooled decision-level metrics on failing inputs:
\[
\mathrm{Recall}=\frac{\sum_i |R_i\cap G_i|}{\sum_i |G_i|},\qquad
\mathrm{Precision}=\frac{\sum_i |R_i\cap G_i|}{\sum_i |R_i|},
\]
with $\mathrm{F1}=2\mathrm{Recall}\mathrm{Precision}/(\mathrm{Recall}+\mathrm{Precision})$ when both are nonzero. Matching is semantic rather than string-based. A constrained diagnostician fills a structured \texttt{gap\_coverage\_analysis} field for each input, and human annotation validates the matches on a stratified sample. Recall measures how completely the model surfaces the decisions the reference program depends on, and precision measures how many of its questions target such decisions.

\subsection{Complexity Proxy and P1 Rate}\label{sec:method:complexity}

\paragraph{Complexity Proxy.}
We use the number of operation tokens in the reference program as a proxy for task complexity, since each additional operation typically corresponds to an additional execution decision fixed by the reference program. For each of the reference programs, we count occurrences of a fixed vocabulary of pandas operations consisting of grouping, merging, joins, sorting, aggregation, filtering, indexing, and comparison, counting each occurrence separately. We split inputs into four tiers by token count.

\paragraph{P1 Rate.}
For model $m$ and tier $t$, the P1 rate is
\[
\mathrm{P1\ Rate}_{m,t}=\frac{|\{q\in t:\mathrm{label}_m(q)=\texttt{Phase1\_Missed\_Gap}\}|}{|\{q\in t\}|}.
\]
The denominator includes every benchmark input prompt in the tier, not only GCG failures, so the rate is the share of all input prompts in the tier attributed to \texttt{Phase1\_Missed\_Gap}, not a share of failures.

\begin{table*}[t!]
\caption{Execution accuracy (\%) of twelve models on BIRD-Pandas under direct generation and under the GCG diagnostic protocol.}
\label{tab:main-comparison}
\centering
\begin{tabularx}{\linewidth}{@{}>{\RaggedRight\arraybackslash}p{0.18\linewidth} *{8}{>{\RaggedLeft\arraybackslash}X}@{}}
\toprule
& \multicolumn{4}{c}{Direct Gen.} & \multicolumn{4}{c}{Grounded Code Generation} \\
\cmidrule(lr){2-5}\cmidrule(lr){6-9}
Model & sim. & mod. & chall. & Over. & sim. & mod. & chall. & Over. \\
\midrule
Llama-3.2-3B & 9.41 & 3.23 & 3.45 & 6.98 & 14.38 & 11.21 & 13.10 & 13.30 \\
Meta-Llama-3-8B & 18.81 & 4.09 & 6.21 & 13.17 & 22.05 & 6.68 & 8.97 & 16.17 \\
Llama-3.2-11B-Vision & 21.30 & 15.30 & 8.97 & 18.32 & 28.86 & 22.20 & 20.69 & 26.08 \\
Qwen3.5-4B & 11.78 & 8.62 & 8.97 & 10.56 & 40.86 & 30.82 & 28.97 & 36.70 \\
Qwen3.5-9B & 24.86 & 25.00 & 16.55 & 24.12 & 44.65 & 38.15 & 33.10 & 41.58 \\
gpt-oss-20b & 50.92 & 39.66 & 38.62 & 46.35 & 57.62 & 45.91 & 41.38 & 52.54 \\
Qwen3.5-27B & 63.68 & 51.94 & 55.86 & 59.39 & 67.35 & 62.28 & 63.45 & 65.45 \\
GPT-5.4-mini & 44.00 & 27.59 & 32.41 & 37.94 & 72.00 & 59.05 & 53.79 & 66.36 \\
Qwen3-Max & 67.03 & 57.97 & 51.03 & 62.78 & 78.59 & 67.03 & 63.45 & 73.66 \\
DeepSeek-V4-Pro & 69.51 & 51.08 & 49.66 & 62.06 & 79.68 & 67.67 & 68.28 & 74.97 \\
Qwen3.6-27B & 65.84 & 50.00 & 53.79 & 59.91 & 82.59 & 71.12 & 68.97 & 77.84 \\
Gemini-3.1-Pro & \textbf{78.81} & \textbf{62.72} & \textbf{62.07} & \textbf{72.36} & \textbf{86.38} & \textbf{74.14} & \textbf{75.86} & \textbf{81.68} \\
\bottomrule
\end{tabularx}
\end{table*}

\section{Experiments}\label{sec:experiments}

\subsection{Experimental Setup}\label{sec:exp:setup}

\paragraph{Dataset.}
We evaluate on BIRD-Pandas~\citep{hu2026bird}, an NL-to-Pandas benchmark of input prompts over 11 databases. The benchmark's official difficulty annotation classifies these into three levels, abbreviated as chall., mod., and sim. in the tables. End-to-end accuracy is computed on all inputs.

\paragraph{Models.}
We evaluate twelve models spanning from 3B to frontier scale. Eight models are run locally on six NVIDIA RTX~A6000 GPUs using each model's official inference code: Qwen3.5-4B, Qwen3.5-9B, Qwen3.5-27B, and Qwen3.6-27B from the Qwen family~\citep{yang2025qwen3,qwenofficialmodels}; Llama-3.2-3B, Llama-3.2-11B-Vision, and Meta-Llama-3-8B from the Llama family~\citep{dubey2024llama3}; and GPT-OSS-20B~\citep{openai2025gptoss}. The remaining four are accessed via API: Qwen3-Max~\citep{qwenofficialmodels}, DeepSeek-V4-Pro~\citep{deepseek2026v4}, Gemini-3.1-Pro~\citep{google2026gemini31pro}, and GPT-5.4-mini~\citep{openai2026gpt54mini}. All models use greedy decoding (temperature~0.0) with fixed random seeds.

\paragraph{Protocol.}
For the evidence phase (Phase 2), we use a reference-conditioned oracle instantiated with Qwen3-Max, except for data queries, which are answered deterministically from the input CSV. To attribute errors, a diagnostic pipeline (Gemini-3.8-flash) processes each failing input prompt and produces a structured attribution along with Phase 1 decision-coverage labels. We use the pipeline-generated labels for all quantitative results.

Following prior work that analyzes failures with LLMs~\citep{ashury2026erroratlas} and checks LLM-produced labels with human annotation~\citep{hu2026bird,clareval2025}, we combine LLM-based analysis with human validation of the labels used for the quantitative results. Three targeted validity checks are reported in Appendix~\ref{app:validation}: (1) human-confirmed precision and recall for the Phase 1 gap coverage analysis on 120 stratified failing input prompts, along with human agreement on the earliest-Phase attribution for 120 failing input prompts drawn from all GCG failures; (2) human annotator agreement on discrepancy labels for 150 input prompts; and (3) human-checked alignment between Phase 1 raised clarification questions and Phase 2 evidence on 120 sampled pairs. These human comparisons serve as validity checks rather than primary results. All reported accuracies are point estimates, and cost analysis is provided in Appendix~\ref{app:cost}.

\subsection{Residual Errors After Answering Raised Questions}\label{sec:exp:main}

To examine whether answering the clarification questions a model raises suffices for correct program synthesis, we compare direct generation with GCG across twelve models at three difficulty levels (Table~\ref{tab:main-comparison}). GCG improves overall accuracy for every model and at every difficulty level, though the magnitude of improvement varies across models. A potential explanation is that some direct-generation failures stem from incorrect implicit assumptions that the clarification answers would correct; this is consistent with but not directly confirmed by the observed variation. Nevertheless, accuracy remains below 100\% in all settings, indicating that residual errors persist even after the raised clarification questions are addressed. Question-level paired results for four representative models are provided in Appendix~\ref{app:paired}.

\begin{table}[t!]
\caption{Share of GCG failures by error attribution category. $N_{\text{diag}}$ is the number of GCG failures on which the diagnostic pipeline produced a label.}
\label{tab:attribution-summary}
\centering
\setlength{\tabcolsep}{3pt}
\begin{tabularx}{\linewidth}{@{}l c c *{3}{>{\Centering\arraybackslash}X}@{}}
\toprule
Model & $N_{\text{diag}}$ & P1 [95\% CI] & P2 & P3-N & P3-T \\
\midrule
Qwen3-Max       & 404 & \textbf{38.6} [34.0, 43.4] & 27.2 & 16.8 & 17.3 \\
DeepSeek-V4-Pro & 384 & \textbf{67.2} [62.3, 71.7] & 9.6  & 6.0  & 17.2 \\
Qwen3.6-27B     & 340 & \textbf{62.4} [57.1, 67.3] & 12.6 & 9.7  & 15.3 \\
Gemini-3.1-Pro  & 281 & \textbf{72.2} [66.7, 77.2] & 13.5 & 2.8  & 11.4 \\
\bottomrule
\end{tabularx}
\end{table}

\subsection{Where Residual Errors Are Lost}\label{sec:exp:isolation}

\paragraph{Error Attribution by Phase.}
To characterize the remaining errors, we assign each GCG failure of the four representative models to the earliest Phase at which the governing execution decision is lost under the protocol (Table~\ref{tab:attribution-summary}). Across all models, \texttt{Phase1\_Missed\_Gap} is the dominant category, accounting for 38.6\% to 72.2\% of failures. Because our priority rule attributes each failure to the phase at which the protocol first loses the decision, each failure is counted only once, at that earliest phase. After the raised clarification questions are answered, the largest share of residual errors is attributed to reference-program execution decisions that were never raised in Phase 1, with a smaller share attributed to insufficient evidence or incorrect program synthesis.

\subsection{What Phase~1 Raises and Misses}

\begin{figure*}[t!]
\centering
\includegraphics[width=0.98\textwidth]{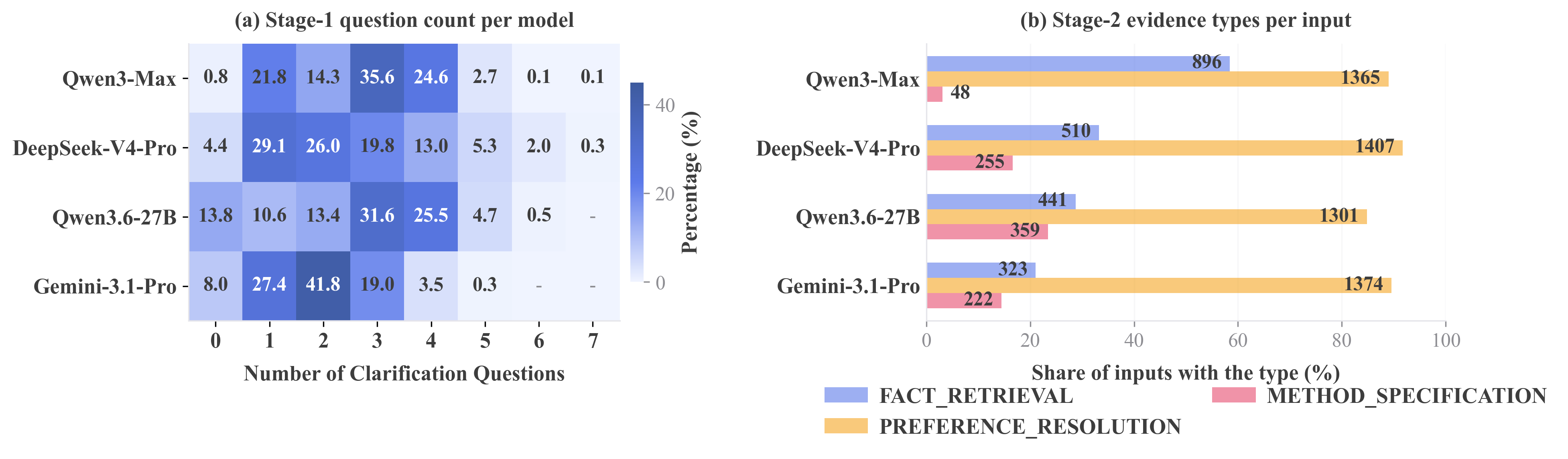}
\caption{Elicitation profile on BIRD-Pandas. (a)~Percentage of input prompts by number of Phase~1 raised clarification questions. (b)~Share (\%) of each model's Phase~2 sub-questions by evidence type.}
\label{fig:elicitation}
\end{figure*}

\paragraph{Number of Raised Questions.}
To understand how many execution decisions models treat as unresolved, we count the Phase 1 raised clarification questions per input prompt (Figure~\ref{fig:elicitation}(a)). For every model, the majority of input prompts receive at least one raised clarification question, with two to three questions covering 45.0\% to 60.8\% of input prompts. This per-prompt question count approximates the number of distinct execution decisions the model identifies as undecidable from the input prompt $\mathcal{I}$ alone, which in turn suggests the extent of implicit assumptions the model makes. Gemini-3.1-Pro concentrates on one to two raised clarification questions per prompt, whereas the other three models more frequently raise three or more.

\paragraph{Evidence Types of Raised Questions.}
To understand what type of information the models seek, we categorize each Phase 2 sub-question by evidence type and report the distribution across types for each model (Figure~\ref{fig:elicitation}(b)). Across all models, \texttt{PREFERENCE\_RESOLUTION} is the predominant type (69.4\% to 78.0\%), followed by \texttt{FACT\_RETRIEVAL} and then \texttt{METHOD\_SPECIFICATION}. This distribution suggests that most clarification questions serve to confirm or correct an assumption the model has already formed, rather than to retrieve a data fact or specify a processing method.

\paragraph{Phase~1 Coverage of Reference Decisions.}\label{sec:exp:clues}

Because this analysis targets residual errors, we compute coverage metrics on the failing input prompts of the four representative models only. Specifically, precision measures the fraction of raised clarification questions that target a reference execution decision, and recall measures the fraction of reference execution decisions covered by a raised clarification question. These metrics characterize coverage among failures and do not generalize to all input prompts.

Precision is consistently high across models, while recall varies more substantially. Models with higher precision tend to have lower recall. To examine how coverage depends on the nature of the input prompt, we also report recall within subgroups of reference execution decisions defined by their Evidence Source and Explicit Level (Table~\ref{tab:coverage}). Recall is lowest when the input prompt provides minimal basis for an execution decision: in every model, recall is lowest for Evidence Source \texttt{N} (no clue in $Q$, $S$, or $K$) and for Explicit Level \texttt{NO\_CLUE}. Although decisions with Evidence Source \texttt{N} constitute a small fraction of all reference execution decisions, most misses involve execution decisions that the input prompt provides partial support for, rather than those with no supporting evidence.

\begin{table*}[t!]
\caption{Phase~1 coverage of reference execution decisions on the failing input prompts.}
\label{tab:coverage}
\centering
\footnotesize
\begin{tabularx}{\textwidth}{@{}>{\RaggedRight\arraybackslash}p{0.15\textwidth} *{10}{>{\Centering\arraybackslash}X}@{}}
\toprule
& & \multicolumn{3}{c}{Phase~1 coverage (\%)} & \multicolumn{4}{c}{Recall by Evidence Source (\%)} & \multicolumn{2}{c}{Recall by Level (\%)} \\
\cmidrule(lr){3-5}\cmidrule(lr){6-9}\cmidrule(lr){10-11}
Model & Decisions & Recall & Prec. & F1 & \texttt{Q} & \texttt{S} & \texttt{K} & \texttt{N} & \texttt{NO\_\allowbreak CLUE} & Other \\
\midrule
Qwen3-Max       & 1{,}128 & 82.4 & 81.6 & 82.0 & 84.0 & 79.5 & 86.8 & 40.9 & 56.4 & 86.0 \\
DeepSeek-V4-Pro & 1{,}050 & 71.0 & 84.7 & 77.3 & 78.5 & 68.1 & 68.7 & 47.5 & 51.3 & 75.5 \\
Qwen3.6-27B     & 1{,}022 & 76.5 & 84.4 & 80.3 & 85.8 & 72.0 & 75.2 & 40.0 & 53.9 & 81.3 \\
Gemini-3.1-Pro  & 642     & 66.5 & 88.8 & 76.0 & 76.2 & 64.9 & 57.8 & 41.4 & 47.5 & 70.8 \\
\midrule
Pooled          & 3{,}842 & 75.1 & 84.2 & 79.4 & 81.5 & 71.9 & 74.8 & 43.0 & 52.5 & 79.5 \\
\bottomrule
\end{tabularx}
\end{table*}

\paragraph{UDC Share by Explicit Level.}

To characterize the unraised execution decisions, we take the execution decisions behind \texttt{Phase1\_Missed\_Gap} failures, which the reference program code depends on and Phase~1 does not raise, and report their distribution over the four Explicit Levels (Table~\ref{tab:missed-gap-level}). In this analysis, UDC denotes missed reference-program decisions whose Explicit Level is not \texttt{EXPLICIT}; these are decisions not stated explicitly in $Q$, $S$, and $K$. This measure accounts for most missed decisions in each model. The decisions are distributed across \texttt{NO\_CLUE}, \texttt{IMPLICIT}, and \texttt{PARTIAL}, indicating that the input provides varying degrees of support for them.

\begin{table}[t!]
\caption{Distribution of identified root-cause gaps by explicit level, for input prompts attributed to \texttt{Phase1\_Missed\_Gap}.}
\label{tab:missed-gap-level}
\centering
\footnotesize
\begin{tabularx}{\linewidth}{@{}>{\RaggedRight\arraybackslash}p{0.23\linewidth} c *{4}{>{\Centering\arraybackslash}X} *{2}{>{\Centering\arraybackslash}X}@{}}
\toprule
Model & $N_{\text{P1}}$ & \texttt{NO\_CLUE} (\%) & \texttt{IMPLICIT} (\%) & \texttt{PARTIAL} (\%) & \texttt{EXPLICIT} (\%) & UDC (\%) \\
\midrule
Qwen3-Max       & 156 & 32.3 & 26.3 & 28.7 & 12.6 & 87.4 \\
DeepSeek-V4-Pro & 258 & 33.3 & 22.3 & 31.1 & 13.3 & 86.7 \\
Qwen3.6-27B     & 212 & 36.1 & 24.9 & 32.2 & 6.8  & 93.2 \\
Gemini-3.1-Pro  & 203 & 32.4 & 19.2 & 36.3 & 12.1 & 87.9 \\
\bottomrule
\end{tabularx}
\end{table}

\subsection{Relation to Program Complexity}

\begin{figure}[t!]
\centering
\includegraphics[width=0.96\linewidth]{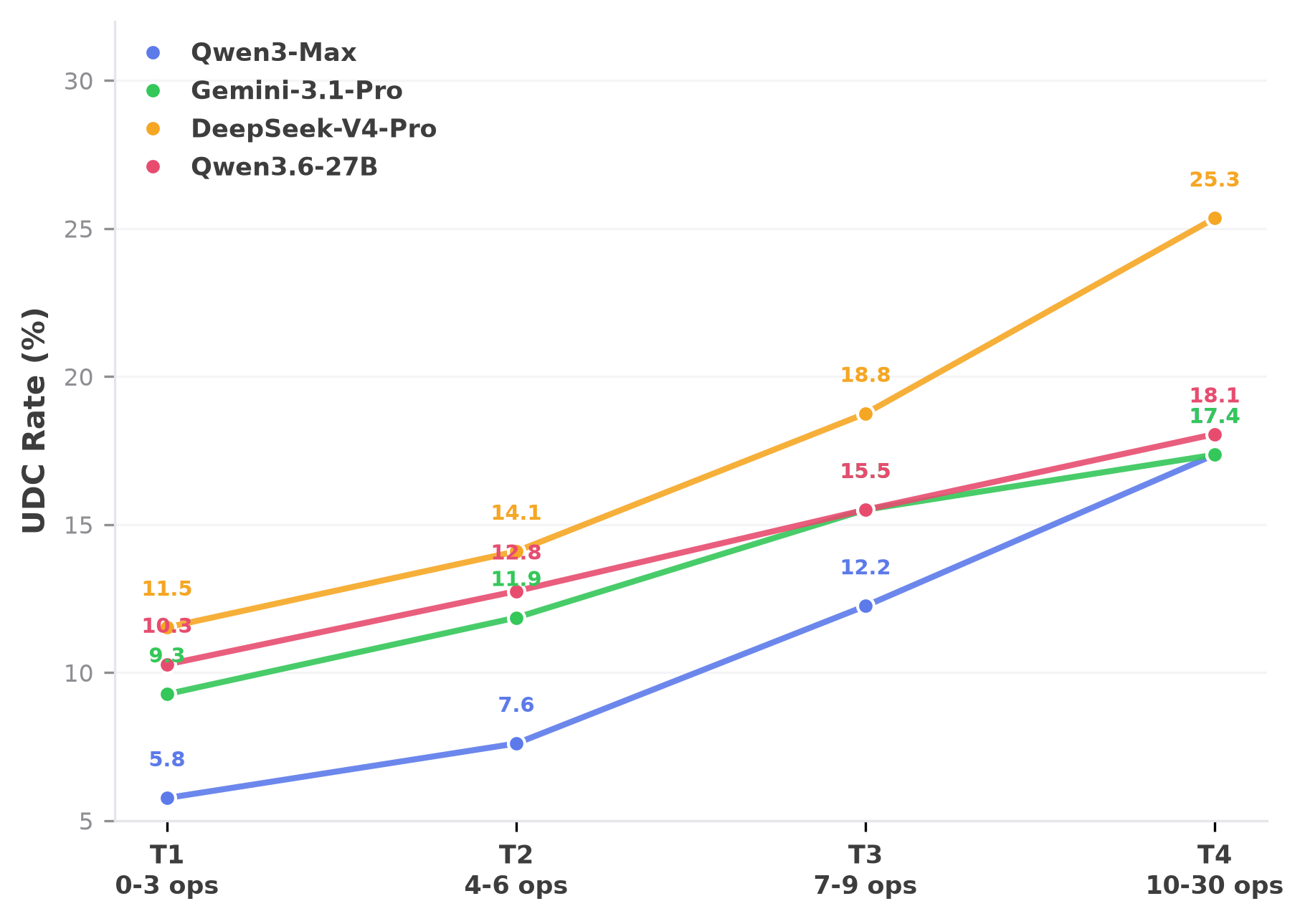}
\caption{\texttt{Phase1\_Missed\_Gap} rate shows a positive association with the reference-program complexity proxy in all four analyzed models.}
\label{fig:ComplexityAssociation}
\end{figure}

\paragraph{Missed-Decision Rate by Complexity Tier.}
To examine how the missed-decision rate varies with program complexity, we group input prompts into four tiers based on the operation-token count of the reference program and report the \texttt{Phase1\_Missed\_Gap} rate within each tier, defined as the share of input prompts in the tier attributed to a missed governing decision (Figure~\ref{fig:ComplexityAssociation}). For all four models, the rate increases monotonically across the four complexity tiers. A longer reference program also encodes more execution decisions for Phase 1 to surface, which may account for the rise of this tier-level rate with program complexity, consistent with a positive association between program complexity and Phase 1 coverage.

\subsection{Testing Alternative Explanations}

\paragraph{Phase~1 Prompt Structure.}
To test whether the residual errors partly stem from insufficient prompt guidance in Phase 1, we compare three settings on three models (Table~\ref{tab:phase1-structure}): direct generation without Phase 1, a free-form Phase 1 prompt, and the full GCG prompt with the five coverage dimensions. Accuracy improves monotonically from direct generation to the free-form prompt and further to the full GCG prompt across all three models. The additional gain from incorporating the five coverage dimensions (9.7 to 18.0 percentage points) exceeds the gain from merely adding a free-form Phase 1 prompt (3.2 to 10.4 percentage points), suggesting that structured coverage dimensions help the model raise execution decisions that unstructured prompting tends to omit.

\begin{table}[t!]
\caption{Execution accuracy (\%) under Phase~1 prompt settings.}
\label{tab:phase1-structure}
\centering
\begin{tabularx}{\linewidth}{@{}>{\RaggedRight\arraybackslash}p{0.32\linewidth} *{3}{>{\Centering\arraybackslash}X}@{}}
\toprule
Model & w/o Phase~1 & + Free-form Phase~1 & + GCG \\
\midrule
GPT-5.4-mini & 37.94 & 48.36 & 66.36 \\
Qwen3.6-27B & 59.91 & 63.87 & 77.84 \\
DeepSeek-V4-Pro & 62.06 & 65.24 & 74.97 \\
\bottomrule
\end{tabularx}
\end{table}

\paragraph{Phase~2 Oracle Strength.}
To test whether residual errors partly result from weak evidence in Phase 2, we fix Phase 1 (using Qwen3.5-27B with the structured prompt) and Phase 3 (Qwen3.5-27B) while varying only the Phase 2 oracle across four models (Table~\ref{tab:oracle}). All oracles use the reference program as the decision-level oracle for rule and method questions, with data questions answered deterministically from the CSV. This controlled comparison isolates the effect of oracle model quality given identical evidence sources. Overall accuracy varies by approximately 3 percentage points across the four oracles, and their relative ranking is not stable across difficulty levels, suggesting that varying the oracle model has a modest impact on end-to-end accuracy under this evidence configuration.

\begin{table}[t!]
\caption{Phase~2 oracle strength ablation.}
\label{tab:oracle}
\centering
\begin{tabularx}{\linewidth}{@{}>{\RaggedRight\arraybackslash}p{0.35\linewidth} *{4}{>{\Centering\arraybackslash}X}@{}}
\toprule
Oracle Model & chall. & mod. & sim. & Over. \\
\midrule
Qwen3-Max & 63.45 & 62.28 & 67.35 & 65.45 \\
DeepSeek-V4-Pro & 63.21 & 61.35 & 66.20 & 64.45 \\
Qwen3.5-27B & 61.66 & 62.52 & 62.40 & 62.37 \\
Qwen3.5-35B-A3B & 63.35 & 58.61 & 63.98 & 62.30 \\
\bottomrule
\end{tabularx}
\end{table}

\section{Limitations and Scope}\label{sec:limitations}

Our diagnostic protocol relies on a reproducible correctness standard. Following BIRD and other Text-to-SQL benchmarks, we use the annotated reference program as the proxy for the intended result~\citep{li2024bird,hu2026bird}, and judge a model's program by executing it alongside the reference program and comparing their outputs. We extend this criterion to clarification: a decision that the reference program depends on counts as covered if Phase 1 raises a question about it. Execution accuracy checks program outputs against the reference, and coverage checks raised decisions against the same reference. This lets us measure which unraised decisions are associated with failures, and it treats the reference program as the working interpretation of each input prompt.

\paragraph{Idealized answers isolate decision elicitation.}
Phase~2 answers come from a reference-conditioned oracle and deterministic CSV queries, so each answer is designed to be consistent with the reference program, and the reference code is withheld from synthesis. This design separates two stages of clarification: whether a decision was raised, and whether its answer was available and correctly used. Appendix~\ref{app:validation:oracle} checks the alignment on the Qwen3-Max error set. The residual errors we measure therefore mainly reflect decisions that Phase~1 did not raise, with evidence quality accounting for a smaller share. This setting fixes the answer source to the reference-conditioned oracle and leaves interactive-use concerns such as retrieval noise, user variability, and multi-turn repair for future work.

\paragraph{Scope of labeling and generalization.}
Attribution and coverage labels are produced by a single automated pipeline, and human checks on a 120-prompt sample, including the earliest-Phase attribution, support the two qualitative patterns the analysis relies on: unraised decisions account for the largest share of residual errors, and among failing prompts, raised questions are mostly relevant. The reported shares are best read as approximate. All results come from single greedy runs on one NL-to-Pandas benchmark in a single-pass setting. In-depth attribution covers four models, and coverage is computed on failing prompts, which matches our focus on residual errors. Complexity is measured with one operation-count proxy and analyzed as a tier-level association with Phase 1 coverage. The findings are descriptive, and extending them to other code-generation tasks or information-access settings requires further validation.

\section{Conclusion}\label{sec:conclusion}

We investigated whether language models can recognize the execution decisions that require clarification before translating natural-language questions into executable programs. Across twelve models, answering clarification questions improved execution accuracy, but incorrect results remained. Our closer examination of four models found that the largest share of these failures arose from decisions the models had never asked about, particularly when the input did not explicitly specify the required choice. This finding points to an important distinction in evaluating natural-language code generation. A model may ask relevant questions and use their answers correctly, yet still fail because it has overlooked other decisions necessary to produce the expected result. By making these omissions observable, GCG provides a way to examine an aspect of model reliability that execution accuracy alone cannot explain.


\bibliographystyle{plainnat}
\bibliography{references}

\appendix

\section{Discrepancy Annotation: Extended Data}\label{app:taxonomy}

\subsection{Object-to-Dimension Mapping}\label{app:taxonomy:mapping}

The primary query object determines the initial Phase~1 dimension. When a discrepancy involves multiple unresolved decisions (e.g., computation may concern business definition, aggregation method, or result cardinality), it maps to multiple dimensions (\texttt{BUSINESS\_RULE}, \texttt{EXECUTION\_METHOD}, or \texttt{RESULT\_CONTRACT}).

\subsection{Annotation Procedure and Category Rules}\label{app:taxonomy:process}

The annotator model (DeepSeek-V4-Pro) compares reference and model programs and returns each logical deviation as a discrepancy with five fields: free-text description, open \texttt{VERB\_OBJECT} label, affected logic, Explicit Level (\texttt{EXPLICIT}/\texttt{PARTIAL}/\texttt{IMPLICIT}/\texttt{NO\_CLUE}), and Evidence Source ($Q$/$S$/$K$/\texttt{N}). A fixed script maps open labels to six failure actions and seven query objects via prefix/keyword rules. Labels come from a single annotator model; Appendix~\ref{app:validation:discrepancy} checks them against human annotation.

\subsection{Representative Discrepancies}\label{app:taxonomy:instances}

Table~\ref{tab:app-gap-tag-examples} lists four representative discrepancies; they span the \texttt{IMPLICIT}, \texttt{PARTIAL}, and \texttt{EXPLICIT} levels and the \texttt{Q}, \texttt{S}, and \texttt{N} evidence sources. A discrepancy qualifies as a Phase~1 candidate when both programs execute, their outputs differ because of the flagged operation, and the input does not fix the choice. For example, QID~590 qualifies (no tie policy stated); QID~22 qualifies only if the schema samples do not reveal that \texttt{sname} contains missing values.

\begin{table*}[t!]
\caption{Representative discrepancies from Qwen3-Max.}
\label{tab:app-gap-tag-examples}
\centering
\footnotesize
\begin{tabularx}{\textwidth}{@{}>{\raggedright\arraybackslash}p{0.14\textwidth} >{\raggedright\arraybackslash}p{0.05\textwidth} >{\raggedright\arraybackslash}p{0.30\textwidth} >{\raggedright\arraybackslash}p{0.19\textwidth} >{\raggedright\arraybackslash}X@{}}
\toprule
\textbf{Query object} & \textbf{QID} & \textbf{Question (excerpt)} & \textbf{Label; level/src} & \textbf{Discrepancy} \\
\midrule
\texttt{COMPUTATION} & 590 & Which user ID has the lowest view? & \texttt{OMIT\_TIE\_HANDLING}; \texttt{IMPLICIT}/\texttt{N} & Model returns first ID (\texttt{idxmin}); reference keeps all tied users. \\
\texttt{RESULT\_CONTRACT} & 71 & What is the district code for the School [\ldots] in Fresno? & \texttt{LIMIT\_TO\_FIRST\_ROW}; \texttt{PARTIAL}/\texttt{Q} & Model returns first code (\texttt{iloc[0]}); reference returns all. \\
\texttt{BOUNDARY\_GUARD} & 22 & Which school in Contra Costa has the highest [\ldots] test takers? & \texttt{OMIT\_SNAME\_NOTNULL}; \texttt{PARTIAL}/\texttt{S} & Model omits \texttt{sname.notna()} filter; may return null. \\
\texttt{SELECTION\_RULE} & 70 & How many active and closed [\ldots] Schools in Alpine? & \texttt{OMIT\_STATUS\_FILTER}; \texttt{EXPLICIT}/\texttt{Q} & Model does not filter by \texttt{StatusType}. \\
\bottomrule
\end{tabularx}
   \end{table*}

The label alone does not decide whether a discrepancy is a Phase~1 candidate; the three conditions above do.

\subsection{Discrepancy Distribution by Model}\label{app:taxonomy:distributions}

Pooled over three models (Qwen3-Max, DeepSeek-V4-Pro, Gemini-3.1-Pro; 2{,}730 discrepancies from 571, 582, and 424 questions), failure actions are \texttt{OMISSION} (44.2\%) and \texttt{WRONG\_SUBSTITUTION} (32.2\%), totaling 76\%. By query object, \texttt{COMPUTATION} is largest (22.7--25.7\%), followed by \texttt{SELECTION\_RULE} (16.5--19.4\%). By Explicit Level, \texttt{NO\_CLUE} dominates (34.7--39.0\%); \texttt{EXPLICIT} is 12.4--18.3\%. By Evidence Source, \texttt{N} (no clue in $Q$/$S$/$K$) is largest (37.9--39.8\%); 60.9\% have a clue. Tables~\ref{tab:app-gap-object} and~\ref{tab:app-gap-clues} detail per-model distributions. \texttt{BOUNDARY\_GUARD} has the highest share with no clue (81.3\%) and \texttt{IMPLICIT}/\texttt{NO\_CLUE} (88.3\%).

\begin{table}[t!]
\caption{Query-object distribution by model.}
\label{tab:app-gap-object}
\centering
\footnotesize
\begin{tabularx}{\linewidth}{@{}>{\RaggedRight\arraybackslash}p{0.30\linewidth} *{5}{>{\Centering\arraybackslash}X}@{}}
\toprule
Query object & Qwen3-Max & DeepSeek-V4-Pro & Gemini-3.1-Pro & No clue (\texttt{N}) & \texttt{IMPLICIT} or \texttt{NO\_CLUE} \\
\midrule
\texttt{COMPUTATION}     & 227 & 244 & 176 & 38.8\% & 52.1\% \\
\texttt{SELECTION\_RULE} & 165 & 181 & 133 & 30.7\% & 50.9\% \\
\texttt{DATA\_REFERENCE} & 145 & 155 & 98  & 32.7\% & 53.5\% \\
\texttt{RESULT\_CONTRACT} & 128 & 131 & 115 & 55.1\% & 66.0\% \\
\texttt{REPRESENTATION}  & 137 & 130 & 64  & 23.3\% & 46.5\% \\
\texttt{RELATIONSHIP}    & 132 & 141 & 57  & 35.8\% & 67.3\% \\
\texttt{BOUNDARY\_GUARD} & 66  & 64  & 41  & 81.3\% & 88.3\% \\
\midrule
Total                    & 1{,}000 & 1{,}046 & 684 & 39.1\% & 57.4\% \\
\bottomrule
\end{tabularx}
\end{table}

\begin{table}[t!]
\caption{Explicit Level and Evidence Source distributions (\%) of the discrepancies of each model (1{,}000, 1{,}046, and 684 discrepancies).}
\label{tab:app-gap-clues}
\centering
\footnotesize
\begin{tabularx}{\linewidth}{@{}>{\RaggedRight\arraybackslash}p{0.17\linewidth} *{8}{>{\Centering\arraybackslash}X}@{}}
\toprule
& \multicolumn{4}{c}{Explicit Level} & \multicolumn{4}{c}{Evidence Source} \\
\cmidrule(lr){2-5}\cmidrule(lr){6-9}
Model & \texttt{NO\_\allowbreak CLUE} & \texttt{IMPL.} & \texttt{PART.} & \texttt{EXPL.} & \texttt{N} & \texttt{Q} & \texttt{S} & \texttt{K} \\
\midrule
Qwen3-Max        & 34.7 & 23.8 & 28.9 & 12.6 & 39.3 & 23.0 & 26.9 & 10.8 \\
DeepSeek-V4-Pro  & 39.0 & 20.1 & 28.5 & 12.4 & 39.8 & 25.8 & 23.9 & 10.5 \\
Gemini-3.1-Pro   & 35.7 & 17.7 & 28.4 & 18.3 & 37.9 & 32.6 & 20.5 & 9.1 \\
\bottomrule
\end{tabularx}
\end{table}

\section{GCG: Design and Illustration}\label{app:framework}

\subsection{Phase~1 Prompt Excerpt}\label{app:framework:Phase1-prompt}

Phase~1 identifies unresolved execution decisions and outputs JSON without generating code. The prompt instructs the model to inspect five coverage dimensions (\texttt{DATA\_REFERENCE}, \texttt{BUSINESS\_RULE}, \texttt{EXECUTION\_METHOD}, \texttt{RESULT\_CONTRACT}, \texttt{BOUNDARY\_GUARD}) and return \texttt{\{"logic\_gaps": [\{"gap\_description": "<assumption, alternatives, rationale>", "clarification\_question": "<which column/rule/method>"\}]\}}. Each gap comprises a \emph{structured triple}: the current assumption, the candidate alternatives, and the rationale explaining why the input ($Q$, $S$, $K$) does not determine the choice.

\subsection{Phase~2 Evidence Routing and Oracle Prompts}\label{app:framework:routing}

Phase~2 decomposes each Phase~1 question into atomic sub-questions and routes them by type: data queries $\to$ \texttt{FACT\_RETRIEVAL} (deterministic CSV queries: \texttt{query\_column} for aggregates/nulls/value-sets, \texttt{query\_filter} for counts/existence, \texttt{query\_cross\_column} for join cardinality); rule verifications $\to$ \texttt{PREFERENCE\_RESOLUTION} (oracle returns \texttt{YES}/\texttt{NO} plus the applicable rule in semantic terms); method clarifications $\to$ \texttt{METHOD\_SPECIFICATION} (oracle returns execution strategy from reference program code). Oracle prompts forbid pandas code and column names, returning only decision-level answers with confidence (\texttt{confirmed}/\texttt{absent}/\texttt{ambiguous}).

\subsection{Phase~3 Prompt Excerpt}\label{app:framework:Phase3-prompt}

Phase~3 receives $\mathcal{H}=(\mathcal{I},\mathcal{C},E,\sigma)$ and generates the pandas program. The prompt provides: (1)~schema $S$ and question $Q$; (2)~evidence preamble binding each type to an action (\texttt{FACT\_RETRIEVAL} $\to$ use data fact; \texttt{PREFERENCE\_RESOLUTION} YES $\to$ add filter, NO $\to$ skip/invert; \texttt{METHOD\_SPECIFICATION} $\to$ translate operation); (3)~serialized evidence $E$ grouped by gap; (4)~external knowledge $K$ for uncovered facts; (5)~constraints (pandas only, no SQL, store result in \texttt{result}, no \texttt{def}/\texttt{lambda}/\texttt{class}). When $E=\emptyset$, the evidence block states ``no logic gaps identified''.

\section{Paired Outcomes}\label{app:paired}

Accuracy totals can hide whether GCG fixes the same questions that direct generation fails on. To check this, we pair the outcomes of direct generation and GCG question by question for the four representative models (Table~\ref{tab:paired-transitions}). For every model, GCG recovers many questions that direct generation gets wrong, and the gain is significant (McNemar test, $p<0.001$). A smaller number of questions change from correct to wrong. One possible reason is that decisions Phase~1 does not raise are still completed silently and can resolve differently once the added questions and answers change the generation context. The paired results show that answering the raised questions reduces errors overall, while some regressions and residual errors remain.

\begin{table}[t!]
\caption{Paired question counts ($N=1{,}534$) from Direct Generation to GCG. C and W mark a correct or wrong answer, so W$\to$C questions are recovered, C$\to$W questions regress, and Net $=$ W$\to$C $-$ C$\to$W.}
\label{tab:paired-transitions}
\centering
\setlength{\tabcolsep}{3pt}
\begin{tabularx}{\linewidth}{@{}l *{5}{>{\Centering\arraybackslash}X}@{}}
\toprule
Model & C$\to$C & W$\to$W & W$\to$C & C$\to$W & Net \\
\midrule
Qwen3-Max       & 865   & 306 & 265 & 98 & +167 \\
DeepSeek-V4-Pro & 885   & 317 & 265 & 67 & +198 \\
Qwen3.6-27B     & 843   & 264 & 351 & 76 & +275 \\
Gemini-3.1-Pro  & 1,043 & 214 & 210 & 67 & +143 \\
\bottomrule
\end{tabularx}
\end{table}

\section{End-to-End Worked Example}\label{app:example-q22}

Table~\ref{tab:q22-gcg} traces Q22, a \texttt{Phase1\_Missed\_Gap} case from the Qwen3.6-27B analysis, through all three Phases under the benchmark's reference-program semantics (Section~\ref{sec:method:preliminaries}). Phase~2 produces evidence for every raised clarification question, although the evidence may be insufficient or misaligned. Phase~3 follows the answers. The remaining error comes from an execution decision the model never raised: the reference program code excludes rows with missing school names via \texttt{sname.notna()}, and the model does not.

\begin{table*}[t!]
\caption{Q22 worked example. Both programs use \texttt{nlargest}; they differ only in the \texttt{sname.notna()} filter.}
\label{tab:q22-gcg}
\centering
\footnotesize
\begin{tabularx}{\textwidth}{@{}p{0.18\textwidth} X@{}}
\toprule
\textbf{Component} & \textbf{Observed content} \\
\midrule
Question & Which school in Contra Costa has the highest number of test takers? \\
Reference program code & \texttt{satscores[(cname == 'Contra Costa') \& (sname.notna())].nlargest(1, 'NumTstTakr')[['sname']]} \\
Phase~1 raised clarification questions & The model asks whether to filter by \texttt{satscores.cname} or join \texttt{schools}; how to resolve ties; which school-name column to display; and how to handle missing \texttt{NumTstTakr}. It does not ask whether the selected row must have a non-null \texttt{sname}. \\
Phase~2 evidence & The raised clarification questions receive reference-consistent answers: use \texttt{cname}, return one row, use \texttt{sname}, and exclude missing \texttt{NumTstTakr}. \\
Phase~3 model program & The model applies the answered conditions and uses \texttt{nlargest(1, 'NumTstTakr')[['sname']]}, but does not apply \texttt{sname.notna()}. \\
Outputs & Reference program code: \texttt{Dougherty Valley High}. Model: \texttt{null}. \\
Earliest loss & \texttt{Phase1\_Missed\_Gap}: the governing execution decision to exclude aggregate rows with missing school names was not covered by the Phase~1 raised clarification questions. \\
\bottomrule
\end{tabularx}
\end{table*}

\section{Validation}\label{app:validation}

\subsection{Human Validation of Phase~1 Gap Coverage Analysis}\label{app:validation:coverage}

The Phase~1 coverage metrics in Section~\ref{sec:exp:clues} are computed from a gap coverage analysis that a constrained LLM diagnostician produces for each failing input prompt. Each entry records an execution decision, whether it appears in the reference program code (\texttt{in\_gold\_code}), whether Phase~1 raised a corresponding clarification question (\texttt{in\_phase1}), and its \texttt{evidence\_source} and \texttt{explicit\_level} labels. Precision and recall are aggregated from these binary judgments.

To check these judgments, two annotators independently label 120 failing input prompts, 30 for each of the four representative models. The prompts are sampled from each model's Phase1\_Missed\_Gap set (156 cases for Qwen3-Max, 258 for DeepSeek-V4-Pro, 212 for Qwen3.6-27B, and 203 for Gemini-3.1-Pro). For each prompt, the annotators compare the Phase~1 clarification questions with the reference program code and label each question as a \emph{semantic match} (addresses the same execution decision as the reference program), a \emph{partial match} (addresses a related but different decision), or a \emph{miss} (addresses no reference decision). They also label each reference execution decision as \emph{identified} (covered by a semantic or partial match) or \emph{missed} (no corresponding Phase~1 question). Cohen's $\kappa$ is 0.81 (88.3\% raw agreement). We recompute precision and recall at the decision level from the human labels and compare them with the pipeline values on the same 120 prompts (Table~\ref{tab:app-coverage-human}). Pooled over the four models, human precision is 86\% versus 85\% for the pipeline, and human recall is 77\% versus 74\%. Human recall is higher than pipeline recall for every model, so the pipeline may slightly underestimate Phase~1 coverage.

We draw a second sample of 120 failing input prompts (30 per model) uniformly from all GCG failures with a diagnostic label. Two annotators see the three-Phase trace and the reference program code, but not the pipeline label, and independently mark the earliest Phase at which the governing execution decision is lost. Cohen's $\kappa$ between annotators is 0.78 (inter-annotator agreement), and the consensus label agrees with the pipeline on 84\% of prompts (101 of 120). Phase1\_Missed\_Gap remains the largest category under both labelings (57\% human consensus versus 60\% pipeline, averaged over the four models). Most disagreements are between Phase1\_Missed\_Gap and Phase2\_Insufficient, where annotators judged that Phase~1 covered the decision but the Phase~2 evidence was insufficient.

\begin{table}[t!]
\caption{Human-confirmed and pipeline Phase~1 gap coverage analysis (\%) on the same 120-prompt sample drawn from each model's Phase1\_Missed\_Gap set. Pipeline values are recomputed on this sample.}
\label{tab:app-coverage-human}
\centering
\footnotesize
\setlength{\tabcolsep}{3pt}
\renewcommand{\arraystretch}{1.15}
\begin{tabularx}{\linewidth}{@{}>{\RaggedRight\arraybackslash}p{0.24\linewidth} *{4}{>{\Centering\arraybackslash}X}@{}}
\toprule
Model & \makecell[c]{Human\\Precision} & \makecell[c]{Pipeline\\Precision} & \makecell[c]{Human\\Recall} & \makecell[c]{Pipeline\\Recall} \\
\midrule
Qwen3-Max       & 86 & 82 & 84 & 82 \\
DeepSeek-V4-Pro & 88 & 85 & 75 & 71 \\
Qwen3.6-27B     & 85 & 84 & 80 & 77 \\
Gemini-3.1-Pro  & 85 & 89 & 70 & 67 \\
\midrule
Pooled (avg)    & 86 & 85 & 77 & 74 \\
\bottomrule
\end{tabularx}
\end{table}

\subsection{Validation of Discrepancy Labels}\label{app:validation:discrepancy}

The discrepancy labels (failure action and primary query object) in Section~\ref{sec:method:taxonomy} come from a single annotator model without human agreement in the main results. Two annotators independently label a stratified sample of 150 input prompts from the three-model pool (50 from each of Qwen3-Max, DeepSeek-V4-Pro, and Gemini-3.1-Pro), assigning the failure action and primary object to each discrepancy the model reports. Cohen's $\kappa$ is 0.79 for the failure action and 0.81 for the primary object, indicating substantial agreement on both axes. Table~\ref{tab:app-discrepancy-human} compares the pooled annotator-model distributions with human consensus. For both failure action and primary object, the two distributions agree within 2 percentage points in every category. The annotator-model labels are a reasonable approximation of human judgment.

\begin{table}[t!]
\caption{Pooled annotator-model and human-consensus distributions (\% of discrepancies) for failure action (left) and primary object (right) on 150 stratified questions.}
\label{tab:app-discrepancy-human}
\centering
\footnotesize
\setlength{\tabcolsep}{3pt}
\renewcommand{\arraystretch}{1.15}
\begin{tabularx}{\linewidth}{@{}>{\RaggedRight\arraybackslash}p{0.36\linewidth} *{4}{>{\Centering\arraybackslash}X}@{}}
\toprule
{} & \multicolumn{2}{c}{Failure action} & \multicolumn{2}{c}{Primary object} \\
\cmidrule(lr){2-3}\cmidrule(lr){4-5}
Category & \makecell[c]{Annotator\\model} & Human & \makecell[c]{Annotator\\model} & Human \\
\midrule
Omission / COMPUTATION           & 44.2 & 45.0 & 23.7 & 24.5 \\
Wrong substitution / SEL.\ RULE  & 32.2 & 31.5 & 17.5 & 18.0 \\
Unwarranted addition / DATA REF & 12.5 & 12.0 & 14.6 & 15.0 \\
Misapplication / RESULT CONTR.   & 5.5  & 6.0  & 13.7 & 13.0 \\
Incorrect computation / REPRES.  & 3.2  & 3.5  & 12.1 & 12.5 \\
Representation mismatch / RELAT. & 2.3  & 2.0  & 12.1 & 11.5 \\
BOUNDARY GUARD                  & ---    & ---  & 6.3  & 5.5  \\
\bottomrule
\end{tabularx}
\end{table}

\subsection{Validation of Phase 2 Evidence Alignment}\label{app:validation:oracle}

\begin{table}[t!]
\caption{Phase 2 evidence alignment.}
\label{tab:app-s2-align}
\centering
\footnotesize
\setlength{\tabcolsep}{3pt}
\renewcommand{\arraystretch}{1.15}
\begin{tabularx}{\linewidth}{@{}l >{\RaggedRight\arraybackslash}X c c@{}}
\toprule
Alignment level & Criterion & Count & Rate (\%) \\
\midrule
Fully aligned     & Evidence precisely answers the semantic need of the question  & 714/973 & 73.4 \\
Partially aligned & Evidence gives the key information but with ambiguity or omission & 187/973 & 19.2 \\
Not aligned       & Evidence is off-target or contradicts the question's direction & 72/973  & 7.4 \\
\midrule
\multicolumn{2}{@{}l}{Fully or partially aligned (LLM judge)}         & 901/973 & 92.6 \\
\multicolumn{2}{@{}l}{Fully or partially aligned (human, $n=120$)}    & 110/120 & 91.7 \\
\bottomrule
\end{tabularx}
\end{table}

\begin{table}[t!]
\caption{Per-question cost of GCG. Every model receives 7{,}264 input tokens. Reasoning tokens are approximate means; times are in seconds.}
\label{tab:app-cost}
\centering
\footnotesize
\begin{tabularx}{\linewidth}{@{}>{\RaggedRight\arraybackslash}p{0.25\linewidth} *{6}{>{\Centering\arraybackslash}X}@{}}
\toprule
Model & \makecell[c]{Reas.\\tokens} & P1 & P2 & P3 & GCG & Direct \\
\midrule
Qwen3-Max       & 4,560 & 127.9 & 13.1 & 120.1 & 261.1 & 121.2 \\
DeepSeek-V4-Pro & 3,200 & 54.3  & 12.4 & 58.7  & 125.4 & 59.1  \\
Qwen3.6-27B     & 3,360 & 50.0  & 12.2 & 52.4  & 114.6 & 52.4  \\
Gemini-3.1-Pro  & 0     & 8.9   & 11.2 & 5.1   & 25.2  & 5.1   \\
\bottomrule
\end{tabularx}
\end{table}

Two annotators independently assess 120 random pairs with Cohen's $\kappa=0.84$. The human fully-or-partially aligned rate is close to the LLM-judged rate (Table~\ref{tab:app-s2-align}), suggesting no large systematic bias.

\section{Clarification Cost and Throughput}\label{app:cost}

Table~\ref{tab:app-cost} reports the model-side cost of clarification; we did not run a user study. For the thinking models, GCG roughly doubles the latency of direct generation, mainly because Phase~1 needs a reasoning pass comparable to Phase~3, while Phase~2 adds a small, nearly constant cost. The non-thinking model adds little in absolute terms. However, Phase~2 is answered by an oracle, so these times exclude human answering time, and the effect of question number and type on usability and throughput in deployment remains untested.

\end{document}